\documentclass[sigconf]{aamas}

\usepackage{balance} % for balancing columns on the final page
\usepackage{booktabs}
\usepackage{multirow}
\usepackage{amsmath}
\usepackage{makecell}
\usepackage{amssymb}
\usepackage{boxedminipage}
\usepackage{listings}
\usepackage{tabularx}
\usepackage{enumitem}
\usepackage{xcolor}
\usepackage[most]{tcolorbox}
\usepackage{fontawesome5}
\usepackage{algorithm}      % 算法环境
\usepackage{algpseudocode}  % 现代伪代码语法
\usepackage{caption}        % 优化标题格式
\usepackage{float}

\newif\ifarxiv
\arxivtrue      % ← arxiv带附录版本
\doi{EXAJ9853}

\makeatletter
\gdef\@copyrightpermission{
  \begin{minipage}{0.2\columnwidth}
   \href{https://creativecommons.org/licenses/by/4.0/}{\includegraphics[width=0.90\textwidth]{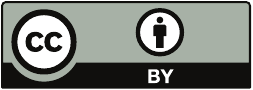}}
  \end{minipage}\hfill
  \begin{minipage}{0.8\columnwidth}
   \href{https://creativecommons.org/licenses/by/4.0/}{This work is licensed under a Creative Commons Attribution International 4.0 License.}
  \end{minipage}
  \vspace{5pt}
}
\makeatother

\setcopyright{ifaamas}
\acmConference[AAMAS '26]{Proc.\@ of the 25th International Conference
on Autonomous Agents and Multiagent Systems (AAMAS 2026)}{May 25 -- 29, 2026}
{Paphos, Cyprus}{C.~Amato, L.~Dennis, V.~Mascardi, J.~Thangarajah (eds.)}
\copyrightyear{2026}
\acmYear{2026}
\acmDOI{}
\acmPrice{}
\acmISBN{}

\fi

\acmSubmissionID{90}

\title[AAMAS-2026 Formatting Instructions]{Parallelized Planning-Acting for Multi-Agent LLM Systems in Minecraft}

\author{Yaoru Li}
\affiliation{
  \institution{Zhejiang University}
  \city{Hangzhou}
  \country{China}}
\email{liyaoru@zju.edu.cn}

\author{Shunyu Liu$^{*}$}
\affiliation{
  \institution{Nanyang Technological University}
  \city{Singapore}
  \country{Singapore}}
\email{shunyu.liu.cs@gmail.com}

\author{Tongya Zheng}
\affiliation{
  \institution{Hangzhou City University}
  \city{Hangzhou}
  \country{China}}
\email{doujiang_zheng@163.com}

\author{Li Sun}
\affiliation{
  \institution{Ningbo Global Innovation Center, Zhejiang University}
  \city{Ningbo}
  \country{China}}
\email{lsun@zju.edu.cn}

\author{Mingli Song}
\affiliation{
  \institution{Zhejiang University}
  \city{Hangzhou}
  \country{China}}
\email{brooksong@zju.edu.cn}

\begin{abstract}
Recent advancements in Large Language Model~(LLM)-based Multi-Agent Systems~(MAS) have demonstrated remarkable potential for tackling complex decision-making tasks. However, existing frameworks inevitably rely on serialized execution paradigms, where agents must complete sequential LLM planning before taking action. This fundamental constraint severely limits real-time responsiveness and adaptation, which is crucial in dynamic environments with ever-changing scenarios like Minecraft.
In this paper, we propose a novel parallelized planning-acting framework for LLM-based MAS, featuring a dual-thread architecture with interruptible execution to enable concurrent planning and acting.
Specifically, our framework comprises two core threads:
(1) a \textit{planning thread} driven by a centralized memory system, maintaining synchronization of environmental states and agent communication to support dynamic decision-making; 
and
(2) an \textit{acting thread} equipped with a comprehensive skill library, enabling automated task execution through recursive decomposition. 
Extensive experiments on Minecraft demonstrate the effectiveness of the proposed framework.
\end{abstract}


\newcommand{\BibTeX}{\rm B\kern-.05em{\sc i\kern-.025em b}\kern-.08em\TeX}

\begin{document}

%%% The following commands remove the headers in your paper. For final 
%%% papers, these will be inserted during the pagination process.

\pagestyle{fancy}
\fancyhead{}

%%% The next command prints the information defined in the preamble.

\maketitle

\begingroup
\renewcommand\thefootnote{\fnsymbol{footnote}}
\footnotetext[1]{Corresponding author.}
\endgroup

% Code available at: \faGithub\ \url{https://github.com/zju-vipa/Odyssey/tree/master/Multi-Agent}

%%%%%%%%%%%%%%%%%%%%%%%%%%%%%%%%%%%%%%%%%%%%%%%%%%%%%%%%%%%%%%%%%%%%%%%%

\section{Introduction}

Multi-Agent Systems (MAS) have become a well-established paradigm for tackling complex decision-making problems~\cite{metagpt, agentverse, villager}, with early efforts primarily relying on reinforcement learning~\cite{marl-survey, marl-survey2, marl1} to enable multiple agents to cooperate or compete in dynamic environments. Despite the encouraging results achieved, these MAS frameworks faced limitations in handling complex real-world scenarios that require advanced communication, reasoning, and adaptability. The rapid advancement of Large Language Models~(LLMs)~\cite{deepseek-r1, deepseek-v3, gpt3, gpt4, qwen2.5, qwen3, llama-3, k1.5} has since revolutionized MAS by adding natural language understanding and generation capabilities, enabling agents to engage in more sophisticated collaboration. LLMs have significantly enhanced the flexibility and versatility of MAS, opening the door to more complex tasks and dynamic interactions in real-world applications~\cite{jubensha,marg,multirobot,sid}.

Recent works have demonstrated the potential of LLM-based MAS in various domains. AgentVerse~\cite{agentverse} improves collaborative performance by orchestrating expert agents, and VillagerAgent~\cite{villager} tackles task dependencies in complex environments using DAG-based task decomposition. Despite these advancements, most current frameworks applied in dynamic environments still rely on serialized execution, where planning and acting occur sequentially for each agent. This serialized nature creates a substantial bottleneck when handling dynamic information, particularly evident in dynamic settings like Minecraft, a game that features a vast and diverse world with various terrains, resources and creatures. Such an environment serves as an ideal testbed for evaluating the capabilities of MAS in open-world scenarios due to its constant environmental changes and rich interaction possibilities. While Voyager~\cite{voyager} pioneered LLM-based agents in Minecraft, it relies on pausing the game server during planning to staticize the environment for the agent in order to prevent interference from environmental changes. However, in real-world dynamic environments, especially in multi-agent systems, it is not feasible to halt the actions of other agents while one agent is planning. This limitation hinders real-time interaction and reduces the system's adaptability to sudden environmental changes or incoming information from other agents during LLM invocations.

Our analysis reveals three critical challenges in current LLM-based MAS for dynamic environments. First, inflexible action scheduling is prevalent, as many existing agent frameworks rely on serialized execution, requiring agents to wait for a language model response before proceeding with further actions. This rigidity complicates the handling of unexpected environmental changes. Second, limited replanning capabilities hinder agents' performance, as they often execute actions to completion without interruption. This lack of adaptability prevents agents from effectively reconsidering or adjusting their plans in response to urgent and unforeseen events, diminishing their overall effectiveness. Lastly, memory sharing delays pose another issue, as memory updates in many multi-agent systems only occur after an action has been fully executed. This results in delayed observational data sharing, causing agents to operate based on outdated information, which in turn limits the team's coordination and efficiency.

In this paper, we propose a parallelized planning-acting framework that introduces a dual-thread architecture with interruptible execution for efficient LLM-based MAS in dynamic environments. Our architecture decouples LLM reasoning from action execution, enabling concurrent planning and acting.  Moreover, the interruption mechanism enables agents to adjust their actions in real time based on environmental changes, thereby improving their adaptability. Specifically, our framework consists of two core threads: (1) A planning thread employing a centralized memory system to support efficient and timely information sharing among agents, minimizing memory sharing delays and ensuring agents operate with up-to-date information for better coordination and efficiency. (2) An acting thread utilizing a comprehensive skill library, enabling efficient task execution through a recursive task decomposition mechanism. Our core contributions are summarized as follows:

\begin{itemize}
\item We propose a parallelized planning-acting framework that decouples planning and acting into dual threads with interruptible execution for efficient LLM-based MAS.
\item We develop a centralized memory system to support the planning thread, ensuring agent decisions are informed by the latest environmental changes and interactions.
\item We design a comprehensive skill library to empower the acting thread, enabling efficient task execution through recursive task decomposition.
\item Experimental results on Minecraft demonstrate a paradigm shift from serialized deliberation to parallelized interaction, yielding notable improvements in efficiency, coordination and adaptability in dynamic environments.
\end{itemize}

\section{LLM-based Multi-Agent Framework}

\begin{figure*}[!t]
\centering
\includegraphics[width=1\textwidth]{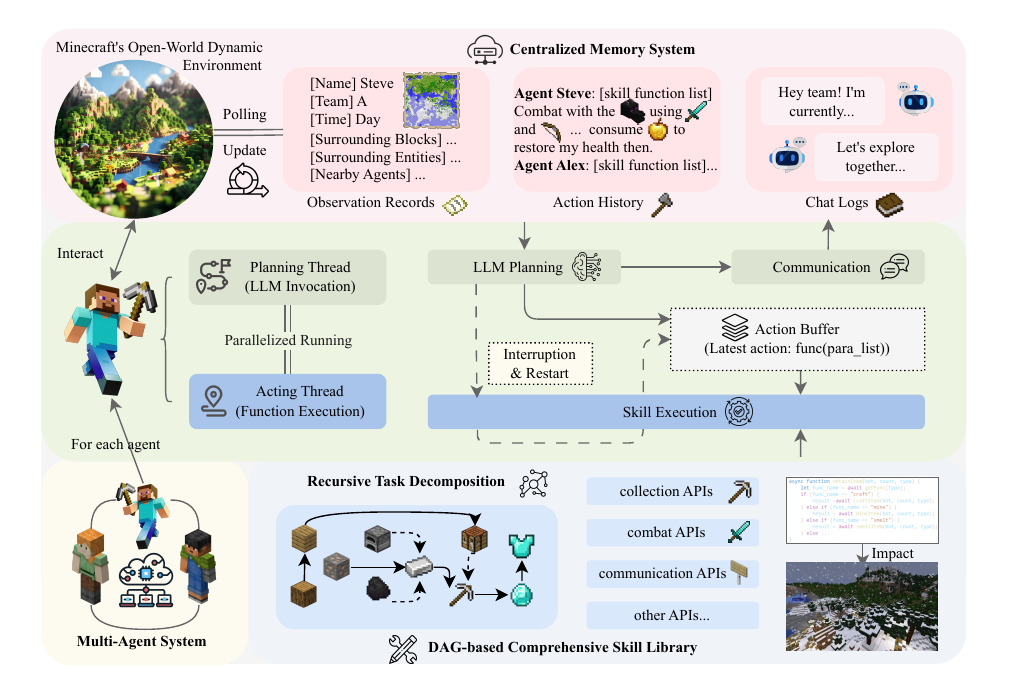}
\caption{An overview of our parallelized planning-acting framework. In the multi-agent system, each agent operates with independent planning and acting threads in parallel. Planning threads are supported by a centralized memory system for decision-making, while acting threads utilize a DAG-based comprehensive skill library for task execution. After each planning step, the latest planned action is stored in the action buffer, and the agent performs a communication. Whenever an action is completed or the interrupt mechanism is triggered, the agent will continue to execute the skill function in the action buffer.}
\label{fig:arch}
\end{figure*}

We propose a novel parallelized planning-acting framework for LLM-based multi-agent systems as shown in Fig.~\ref{fig:arch}, designed to support real-time inter-agent collaboration in dynamic scenarios such as the open-world environment of Minecraft. Our framework introduces three key innovations: (1) A dual-thread architecture with an interruptible execution mechanism, enabling concurrent planning and acting, (2) A real-time updated centralized memory system supporting the planning thread, ensuring that agents' decisions are informed by the latest environmental changes and team communications, and (3) A comprehensive skill library supporting the acting thread, automating task execution by proposing a recursive task decomposition mechanism.

\subsection{Parallelized Planning-Acting Framework}

Inspired by the human ability to think and act simultaneously, our framework adopts a dual-thread architecture (Fig.~\ref{fig:arch}) that decouples planning (driven by LLMs and the centralized memory system) from acting (executed by a comprehensive skill library). Let $ \mathcal{G} = \{g_1, g_2, \dots, g_n\} $ denote the set of agents. The planning and acting threads operate asynchronously and independently, communicating only through a shared action buffer.

\begin{itemize}
\item \textbf{Planning Thread}:
The planning thread continuously monitors the environment and generates new action proposals for agent $ g_i $ based on the system prompt $ S $, the agent's current observation $ O_i $, the latest team chat logs $ C $, and its current action $ A_i $. At any time, the LLM may propose a new action together with an interruption flag:

\begin{equation}
A_i^{\text{new}}, flag_{intr} = \text{LLM}(S, O_i, C, A_i).
\end{equation}

This proposed action $ A_i^{\text{new}} $ is then written into a shared \textit{action buffer}, which acts as a communication channel between the planning and acting threads. The buffer is implemented as a single-slot queue: if it is already occupied, the previous action will be overwritten. This ensures that the buffer always holds the most up-to-date action recommendation from the planner, reflecting the latest context. The interruption mechanism is also completely controlled by the LLM: it may trigger a restart if the new action is judged more urgent, or if the current action is no longer meaningful. When $flag_{intr}=\text{True}$, the planner issues a restart signal to the acting thread, ensuring that ongoing execution will be terminated and replaced by the new plan.

\item \textbf{Acting Thread}:  
The acting thread is responsible for executing skills from the comprehensive skill library. Let $ A_i^{\text{curr}} $ denote the action currently being executed, and $ A_i^{\text{new}} $ the action in the buffer. Upon restart, the acting thread immediately aborts its ongoing skill execution and fetches the latest action from the buffer. Otherwise, it continues executing $ A_i^{\text{curr}} $ until completion, while periodically checking for updates. This design makes the acting thread reactive to planner-issued interrupts, rather than making its own decisions about preemption.
\end{itemize}

This decoupled design allows flexible, interruptible execution: planners can revise intentions at high frequency, while actors respond dynamically by replacing or preempting ongoing actions. Compared to fixed scheduling, this architecture significantly enhances responsiveness and adaptability in dynamic environments. Algorithm~\ref{alg:parallel} summarizes the procedure.

\begin{algorithm}[htbp]
\caption{Parallelized Planning-Acting Framework}
\label{alg:parallel}
\begin{algorithmic}[1]
\State Initialize agent set $\mathcal{G} \gets \{g_1, \dots, g_n\}$
\State $M \gets$ Centralized Memory System
\State $Q \gets \text{Queue}(1)$ \Comment{Single-slot action buffer}

\Procedure{PlanningThread}{$g_i$}
\While{not $\text{TaskCompleted}()$}
    \State \texttt{O\_i} $\gets$ \texttt{GetObservation}(\texttt{g\_i})
    \State \texttt{C} $\gets$ \texttt{M.GetChatLogs}()
    \State (\texttt{new\_action}, \texttt{intr}) $\gets$ \texttt{LLM}(S, \texttt{O\_i}, \texttt{C}, \texttt{cur\_action})
    \If{$\texttt{intr} = True$}
        \State \texttt{Restart(ActingThread($g_i$))}
    \EndIf
    \State $Q.\text{put}(\texttt{new\_action}, \text{overwrite=True})$
\EndWhile
\EndProcedure

\Procedure{ActingThread}{$g_i$}
\While{not $\text{TaskCompleted}()$}
    \If{$Q.\text{has\_new}()$}
        \State \texttt{new\_action} $\gets Q.\text{get}()$
        \State \texttt{cur\_action} $\gets$ \texttt{new\_action}
    \EndIf
    \If{$\texttt{cur\_action} \neq \text{None}$}
        \State \texttt{ExecuteSkill(cur\_action)}
        \State $M.\text{UpdateExecution}(\texttt{cur\_action})$
    \EndIf
\EndWhile
\EndProcedure

\State Start \textsc{PlanningThread} and \textsc{ActingThread} for each $g_i \in \mathcal{G}$
\State Wait until all threads terminate
\end{algorithmic}
\end{algorithm}

\textbf{Latency Analysis.}
The parallelized architecture intuitively reduces system latency through concurrent execution of planning and acting threads. Let $T_{\text{plan}}$ denote the LLM reasoning latency and $T_{\text{act}}$ the skill execution time. For a task requiring $n$ atomic actions without any interruption:

\begin{itemize}
    \item \textbf{Serialized Framework:} 
    \begin{equation}
        T_{\text{s}} = \sum_{k=1}^n (T_{\text{plan}}^{(k)} + T_{\text{act}}^{(k)}).
    \end{equation}
    
    \item \textbf{Parallelized Framework:}
\begin{equation}
    T_{\text{p}} = T_{\text{plan}}^{(1)} + \sum_{k=2}^n \max(T_{\text{plan}}^{(k)}, T_{\text{act}}^{(k-1)}) + T_{\text{act}}^{(n)}.
\end{equation}
\end{itemize}

The latency reduction $\Delta T$ can be expressed as:
\begin{equation}
    \begin{split}
        \Delta T &\approx \sum_{k}^n \left( T_{\text{plan}}^{(k)} + T_{\text{act}}^{(k)} - \max\left(T_{\text{plan}}^{(k)}, T_{\text{act}}^{(k)}\right) \right) \\
        &= \sum_{k}^n T_{\text{plan}}^{(k)} \quad \text{if for all } k, \, T_{\text{act}}^{(k)} > T_{\text{plan}}^{(k)}
    \end{split}
\end{equation}

This analysis highlights that the overlapping of planning and acting phases successfully conceals $T_{\text{plan}}$ when $T_{\text{act}} > T_{\text{plan}}$. We propose the comprehensive skill library in Section~\ref{sec:skill} to ensure this condition is well-maintained. For instance, a complex long-duration skill that might take several minutes to execute can be recursively decomposed and automated, while LLM reasoning typically only requires a few seconds. 
It is important to emphasize that our primary goal is to enable real-time and dynamic interaction, while latency reduction is a secondary benefit. 
Fig.~\ref{fig:timeline} vividly presents a compact timeline diagram that illustrates the overlap between planning and acting, along with a concrete interrupt example.

\subsection{Centralized Memory System}

To facilitate effective coordination, we implement a centralized memory system \( M \) that stores and manages information at the team level. The memory is updated at each time step \( t \) as follows:
\begin{equation}
M^{t+1} = \{O^{t+1}, C^{t+1}, A^{t+1}\} \cup \{M^t \setminus O^t\},
\end{equation}
where \( O^{t+1} \) denotes the updated observations of the multi-agent system at time \( t+1 \), which overwrite the previous observations \( O^t \), \( C^t \) denotes the chat messages of the system at time \( t \), \( A^t \) denotes the action history of the system at time \( t \). 
This unified repository enables agents to access and utilize relevant information during task execution, ensuring efficient team coordination:

\begin{figure}[!h]
\centering
\includegraphics[width=0.5\textwidth]{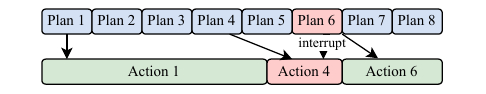}
\caption{Schematic diagram of the timeline for parallel operation of the planning-acting dual threads. Among them, plans 2-3 are deemed not urgent enough to trigger the interrupt mechanism and thus are not actually executed. After the action 1 corresponding to plan 1 is completed, the latest action 4 corresponding to plan 4 is executed directly. Plan 6 sends an interrupt signal to suspend action 4, and the execution of action 6 starts immediately.}
\label{fig:timeline}
\end{figure}

\begin{itemize}
    \item \textbf{Observation Records}:  
    Each agent's observations are continuously updated in a polling manner (\textit{e.g.} update per second), reflecting the latest agent status and environmental state. These observations are associated with the respective agent, allowing the team to maintain a comprehensive and up-to-date view of the environment.

    \item \textbf{Chat Logs}:  
    Team chat messages are always updated in real time, with long-term retention to support historical analysis and decision making. During planning, agents can retrieve the most recent messages to incorporate team insights into their strategies, ensuring team collaboration.

    \item \textbf{Action History}:  
    Actions taken by each agent are also recorded, providing a detailed history of task execution. During planning, agents need to make decisions based on their current action and determine whether to interrupt it.
\end{itemize}

We implement two types of multi-agent communication, including passive communication and active communication, ensuring a dynamic and diverse resource for team coordination among agents:

\begin{itemize}
    \item \textbf{Passive Communication}:  
    In the planning thread, the LLM generates a chat message based on the agent's latest observations after planning, which is then sent to the centralized memory's chat logs. This ensures that updated observations can run concurrently with action execution. While an agent is performing actions, its observations are continuously updated and shared with the team, enabling real-time coordination based on the most current environmental information.

    \item \textbf{Active Communication}:  
    In the acting thread, agents can actively choose to send chat messages by performing a \texttt{chat} action implemented by the comprehensive skill library. This allows the agent to share any information with teammates, updating the chat logs in real-time. This form of communication ensures that agents can respond dynamically and share critical information, rather than passively communicating after the current action is over.
\end{itemize}

By supporting both passive and active communication modes, this system effectively addresses the challenge of memory sharing latency, ensuring that agents always operate on the latest team knowledge. For concrete examples illustrating these concepts, please refer to Appendix.

\subsection{Comprehensive Skill Library}
\label{sec:skill}
To enable seamless interaction between agents and the Minecraft environment, we develop a comprehensive skill library based on Mineflayer~\cite{Mineflayer} that encapsulates a wide range of in-game actions. The library provides high-level APIs for tasks such as resource collection, combat, exploration, and communication. For further technical details, please refer to Appendix.

Our comprehensive skill library implements a recursive task decomposition mechanism, which automates the completion of prerequisite tasks such as mining raw materials and crafting necessary tools. This automation ensures that agents can perform complex resource collection tasks with minimal manual intervention, enabling the automated collection of over 790 types of items in Minecraft, surpassing all existing methods~\cite{voyager, gitm, steve, odyssey}.

The core recursive process can be formally modeled as a weighted directed acyclic graph (DAG) $\mathcal{G} = (V, E, \phi)$, where:
\begin{itemize}
\item \textbf{Vertex set} $V = \{v_i\}$ represents atomic tasks:
\begin{equation} v_i = (t_i, c_i, f_i), \end{equation}
where $t_i \in \mathcal{T}$ is the target item type (All collectible items in Minecraft), $c_i \in \mathbb{N}^+$ denotes required quantity, and $f_i$ specifies the operation type for obtaining the item.
\item \textbf{Edge set} $E \subseteq V \times V$ encodes task dependencies:
\begin{equation} (v_j, v_i) \in E \iff v_j \in \text{pre}(v_i), \end{equation}
where $\text{pre}(v_i)$ gives the prerequisite tasks of $t_i$.
\item \textbf{Weight function} $\phi: E \to \mathbb{Q}^+$ defines conversion rates:
\begin{equation}
\phi(v_j, v_i) = \frac{r_{ij}}{n_{\text{out}}},   
\end{equation}
with $r_{ij}$ being the required quantity of $t_j$ and $n_{\text{out}}$ being the output quantity per operation.
\end{itemize}

The recursive resolution process follows:
\begin{equation}
\Psi(v_i) = \bigcup_{(v_j, v_i) \in E} \left\{\Psi\left(v_j^{\left(\phi(v_j, v_i) \cdot c_i\right)}\right)\right\} \cup \{v_i\},
\end{equation}
where $v_j^{(k)}$ denotes a task requiring $k$ units of $t_j$, and the base case $\Psi(v_i) = \emptyset$ applies when $I(t_i) \geq c_i$, with $I$ representing the current inventory state, which means the task is accomplished.

This recursive task decomposition mechanism effectively models task prerequisite relationships as a Directed Acyclic Graph (DAG), where complex tasks are dynamically decomposed into atomic subtasks through automated dependency resolution. Unlike conventional approaches that require explicit step-by-step navigation from initial states to target objectives, our implementation enables agents to directly invoke high-level skills while the system automatically handles the recursive resolution of all prerequisite conditions. This design allows our multi-agent system to remain efficient by offloading detailed task execution to the skill library while leveraging LLMs for strategic decision-making and dynamic prioritization.

Moreover, the recursive task decomposition mechanism can be generalized to other environments beyond Minecraft. It is intuitive that allowing an LLM to independently plan the task dependencies embedded within a DAG would not be as efficient as utilizing a fixed mechanism. Consider a Directed Acyclic Graph (DAG) $\mathcal{G} = (V, E)$ with $n = |V|$ nodes and $e = |E|$ edges, representing task dependencies. Let the shortest path length from the source node $v_s$ (a node with zero in-degree) to a target node $v_t$ (a node with zero out-degree) be denoted as $L(v_s, v_t)$. This path corresponds to the minimal sequence of prerequisite tasks that must be executed to achieve the final objective represented by $v_t$.

In traditional planning approaches where each task is sequentially inferred by an LLM, the number of required model calls is at least $L(v_s, v_t) + 1$, i.e., one for each node along the critical path. In contrast, our mechanism resolves all prerequisite dependencies automatically through the DAG structure, requiring only a single LLM invocation for the high-level task $v_t$.

This significant reduction in LLM usage not only improves computational efficiency but also reduces potential error propagation across multiple reasoning steps. Therefore, our methodology offers a generalizable framework for integrating LLMs with structured task execution systems across diverse domains beyond Minecraft.

The comprehensive skill library also demonstrates strong scalability and can be readily extended to accommodate new updates. Newly introduced skills can be seamlessly integrated by simply updating the prerequisite dependency DAG that encodes task relationships without new interfaces development or extensive code modifications. This modular design ensures continued functionality with minimal maintenance effort, making the library both future-proof and adaptable to evolving domain requirements.

\section{Experiments}
\label{sec:exp}
\subsection{Benchmark Task Design}
\label{sec:exp_design}

Our experiments are designed to validate the framework's capabilities, leveraging Minecraft as a testbed while focusing on general Multi-Agent System (MAS) competencies. Existing Minecraft agent methods suffer from a lack of a comprehensive skill library, which hinders their ability to learn complex strategies for tackling challenging tasks. Consequently, we have developed more demanding tasks built upon the existing evaluation paradigm, rather than limiting our scope to basic tasks solely for baseline comparisons. For example, defeating the Ender Dragon is widely recognized as Minecraft's ultimate challenge, which remains unachievable with current methods to the best of our knowledge, serves as one of our evaluation tasks. Our benchmark comprises a diverse range of tasks and supports standardized evaluation for potential future research. Refer to the Appendix for more details.

\begin{itemize}
\item \textbf{Resource Collection Task:} Evaluates the foundation of our framework through fundamental mining and crafting tasks in Minecraft. These tasks aim to validate the effectiveness of our comprehensive skill library by requiring agents to complete compound items with deep dependency chains and measure multi-agent coordination efficiency in distributed resource collection. we define a set of representative resource collection tasks, such as \textit{Diamond Armor} serves as a foundation for challenging combat scenarios in the game.

\item \textbf{Boss Combat Task:} Assesses dynamic adaptation and strategical coordination through combat scenarios against powerful bosses. In Minecraft, there are three primary world dimensions: Overworld, Nether, and End. Each dimension hosts powerful boss monsters whose defeat is considered pinnacle challenges for players of Minecraft. Based on this, we define three representative combat tasks.

\item \textbf{Adversarial PVP Task:} Compares two frameworks directly by having them engage in direct combat scenarios. In this task, two teams of agents engage in battles with each other. 

\end{itemize}

\subsection{Experimental Setup}

All experiments are conducted in the Minecraft gaming environment using the Qwen-Plus model~\cite{qwen2.5}, while multi-modal experiments utilized the Qwen-VL-Plus model~\cite{qwen-vl}. The game server operates continuously without pausing during interactions with LLMs, so all agents perform in real time. See Appendix for more details.

\subsection{Main Results}

\subsubsection{Resource Collection Task}
\label{sec:resource}

We evaluate our framework's performance on eight composite resource collection tasks in Minecraft. Each task requires agents to collect multiple items, where each item involves a long chain of prerequisite dependencies. For instance, crafting \textit{Diamond Armor} necessitates first gathering wood, stone, and iron to produce the required tools, and all of which can be automatically accomplished through the recursive decomposition of our skill library. Agents only need to collaborate at the skill level. Table~\ref{tab:resource_collection} compares the completion times between our multi-agent system and the single-agent baseline, demonstrating the efficiency gains achieved through coordination.

\begin{table}[htbp]
\centering
\caption{Average completion time and standard deviation of resource collection tasks comparing multi-agent (MA, fixed at 3 agents) and single-agent (SA) systems over 10 trials.}
\label{tab:resource_collection}
\begin{tabular}{ccc}
\toprule
\makecell[c]{\textbf{Task}} & \textbf{MA Time (min)} & \textbf{SA Time (min)} \\ \midrule
\makecell[c]{Iron Tool Set\\ \includegraphics[width=0.5cm]{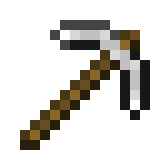} \includegraphics[width=0.5cm]{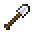} \includegraphics[width=0.5cm]{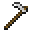}\includegraphics[width=0.5cm]{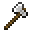}} 
& $ 7.8 \pm 2.1 $ & $ 8.5 \pm 3.7 $ \\
\makecell[c]{Diamond Armor\\ \includegraphics[width=0.5cm]{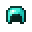}\includegraphics[width=0.5cm]{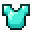} \includegraphics[width=0.5cm]{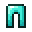} \includegraphics[width=0.5cm]{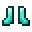}} 
& $ 13.7 \pm 4.1 $ & $ 28.3 \pm 6.1 $ \\
\makecell[c]{Redstone Devices\\ \includegraphics[width=0.5cm]{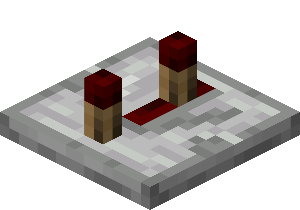} \includegraphics[width=0.5cm]{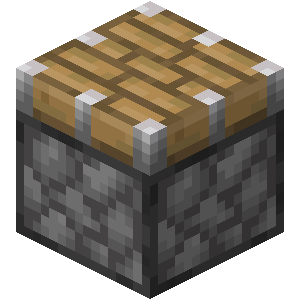} \includegraphics[width=0.5cm]{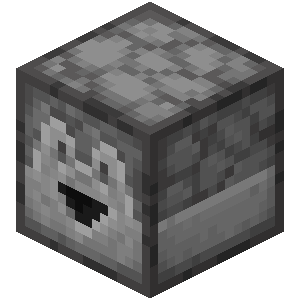}} 
& $ 11.0 \pm 6.0 $ & $ 13.1 \pm 3.3 $  \\
\makecell[c]{Navigation Kit\\ \includegraphics[width=0.5cm]{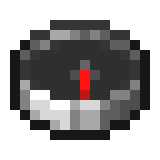} \includegraphics[width=0.5cm]{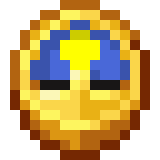} \includegraphics[width=0.5cm]{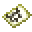}} 
& $ 25.3 \pm 12.2 $ & $ 39.4 \pm 11.7 $ \\
\makecell[c]{Transport System\\ \includegraphics[width=0.5cm]{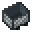} \includegraphics[width=0.5cm]{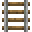} \includegraphics[width=0.5cm]{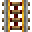}} 
& $ 22.0 \pm 10.1 $ & $ 37.8 \pm 12.6 $ \\
\makecell[c]{Food Supplies\\ \includegraphics[width=0.5cm]{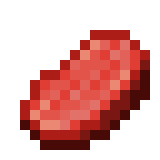} \includegraphics[width=0.5cm]{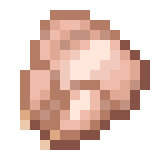} \includegraphics[width=0.5cm]{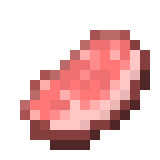}} 
& $ 6.6 \pm 3.9 $ & $ 8.0 \pm 2.0 $ \\
\makecell[c]{Building Materials\\ \includegraphics[width=0.5cm]{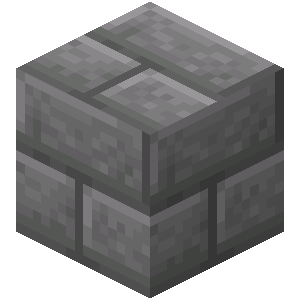} \includegraphics[width=0.5cm]{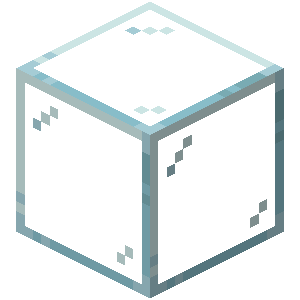} \includegraphics[width=0.5cm]{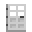}} 
& $ 15.8 \pm 2.8 $ & $ 22.6 \pm 7.4 $ \\
\makecell[c]{Storage System\\ \includegraphics[width=0.5cm]{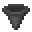} \includegraphics[width=0.5cm]{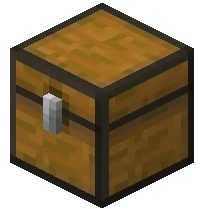} \includegraphics[width=0.5cm]{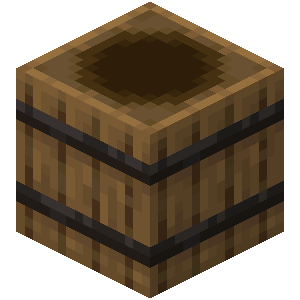}} 
& $ 10.0 \pm 8.9 $ & $ 16.7 \pm 7.8 $ \\
\bottomrule
\end{tabular}
\end{table}

The experimental results demonstrate that our Minecraft comprehensive skill library, when employed within a multi-agent framework, can efficiently complete various resource collection tasks. As shown in the comparisons, the system with three agents significantly reduces task completion times compared to the single-agent baseline across most tasks, validating the effectiveness of our approach and showcasing the efficiency benefits of multi-agent collaboration. However, the performance gain does not scale linearly with the number of agents, which mainly stems from sequential dependency chains and spatial contention.

%In practice, the total task completion time remains higher than the theoretical lower bound assuming perfect division of labor among all agents. This suggests that the expected proportional speedup is not consistently realized, which is due to two inherent limitations:

%\begin{itemize}
%    \item Resource dependency chains: Sequential prerequisites in  workflows (\textit{e.g.}, iron tools required for diamond mining)
%    \item Spatial contention: Overlapping access to resources in close proximity (\textit{e.g.}, multiple agents competing to gather items from the same place)
%\end{itemize}

\subsubsection{Boss Combat Task}
We conduct comprehensive evaluations across three challenging combat scenarios, each involving different agent team sizes and featuring an extremely powerful and representative boss in Minecraft: \textit{Elder Guardian}, \textit{Wither}, and \textit{Ender Dragon}. Each task requires agents to dynamically adjust their strategies based on environmental changes. For instance, when fighting the \textit{Wither}, agents must monitor whether it has entered its final phase where it becomes immune to ranged attacks; when battling the \textit{Ender Dragon}, agents need to first destroy the end crystals that continuously restore its health. Prior to each combat scenario, we equip each agent with identical initial supplies, including weapons, armor, and consumables. Our proposed comprehensive skill library supports the collection and crafting of these supplies, for example, the crafting of \textit{Diamond Armor} has already been validated in previous experiments. We also observe that although each agent is homogeneous at initialization, heterogeneous division of labor can emerge through multiple rounds of actions and communication within the multi-agent system.

The performance of the multi-agent system in the boss combat task is summarized in Table~\ref{tab:boss_combat}, demonstrating that our framework achieves high success rates in completing all challenging boss combat tasks with various agent team sizes.

\begin{table*}[htbp]
    \centering
    \caption{Performance of boss combat across three task scenarios, reporting mean values and standard deviations of multiple evaluation metrics over 12 trials. "\#Agents" refers to the number of agents. "Time" refers to the minutes taken to complete the combat and achieve victory and is calculated only for successful trials. "Health Ratio" refers to the ratio of the remaining health value of the team to its full health. 'Progress' refers to the percentage of damage dealt to the boss to its full health. See Appendix for detailed metric definitions.}
    \label{tab:boss_combat}
    \begin{tabular*}{\textwidth}{@{\extracolsep{\fill}}cccccc}
        \toprule
        \textbf{Task Scenario} & \textbf{\# Agents} & \textbf{Time~(min)} & \textbf{Health Ratio} & \textbf{Progress} & \textbf{Success Rate} \\ 
        \midrule
        \multirow{3}{*}{\makecell{Elder Guardian (Overworld)\\ \includegraphics[width=1.2cm]{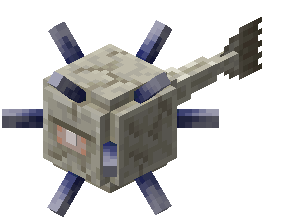}}}
        & 3 & $2.4 \pm 2.1$ & $49.8 \pm 32.8\%$ & $91.4 \pm 20.1\%$ & $83.3\%$ \\ 
        & 5 & $1.2 \pm 0.8$ & $84.4 \pm 7.9\%$ & $100.0 \pm 0.0\%$ & $100.0\%$ \\ 
        & 10 & $1.2 \pm 0.3$ & $86.9 \pm 8.4\%$ & $100.0 \pm 0.0\%$ & $100.0\%$ \\
        & 20 & $1.1 \pm 0.4$ & $89.3 \pm 13.4\%$ & $100.0 \pm 0.0\%$ & $100.0\%$ \\ 
        \midrule
        \multirow{3}{*}{\makecell{Wither (the Nether)\\ \includegraphics[width=1.2cm]{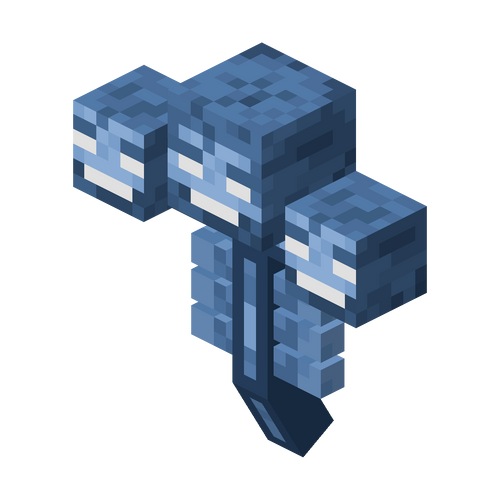}}}
        & 3 & $1.8 \pm 0.8$ & $18.6 \pm 23.4\%$ & $71.8 \pm 35.9\%$ & $41.7\%$ \\ 
        & 5 & $1.5 \pm 0.5$ & $53.4 \pm 32.7\%$ & $88.1 \pm 20.6\%$ & $75.0\%$ \\ 
        & 10 & $1.4 \pm 0.3$ & $69.5 \pm 19.0\%$ & $100.0 \pm 0.0\%$ & $100.0\%$ \\ 
        & 20 & $1.0 \pm 0.1$ & $73.4 \pm 3.9\%$ & $100.0 \pm 0.0\%$ & $100.0\%$ \\
        \midrule
        \multirow{3}{*}{\makecell{Ender Dragon (the End)\\ \includegraphics[width=1.2cm]{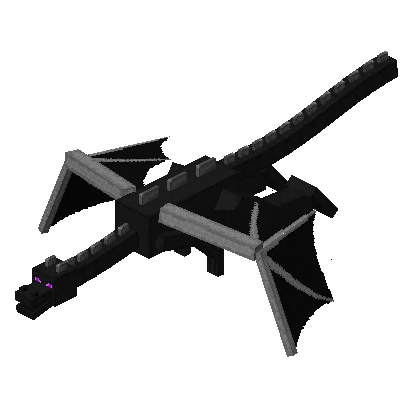}}}
        & 3 & $ N/A $ & $0.0 \pm 0.0\%$ & $20.4 \pm 18.5\%$ & $0.0\%$ \\ 
        & 5 & $6.5 \pm 2.0$ & $18.1 \pm 23.2\%$ & $75.4 \pm 28.9\%$ & $41.7\%$ \\ 
        & 10 & $5.2 \pm 1.5$ & $43.9 \pm 24.4\%$ & $98.2 \pm 5.9\%$ & $91.7\%$ \\
        & 20 & $2.9 \pm 0.8$ & $67.9 \pm 13.6\%$ & $100.0 \pm 0.0\%$ & $100.0\%$ \\ 
        \bottomrule
    \end{tabular*}
\end{table*}

\subsubsection{Adversarial PVP Task}

\begin{figure}[!t]
\centering
\includegraphics[width=0.5\textwidth]{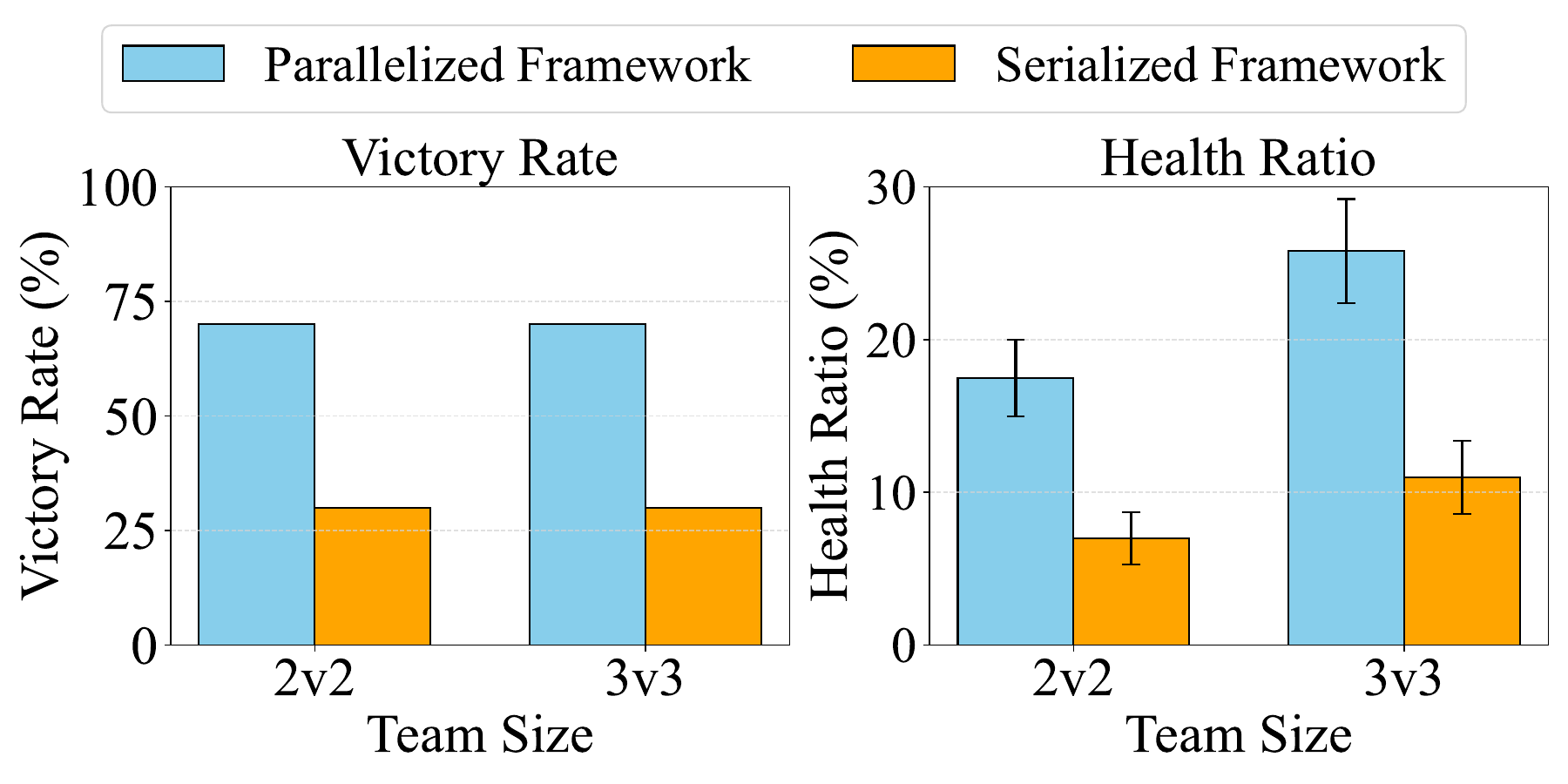}

\caption{Comparison of parallelized vs. serialized frameworks in PVP tasks. All results are shown with the mean and standard deviation over 10 trials.}

\label{fig:pvp}
\end{figure}

 Similar to the Boss Combat Task, in Adversarial PVP Task, agents were provided with initial combat resources, and then the two teams of agents engaged in battles with each other. We conduct a direct comparison experiment between the parallelized and serialized frameworks across various team sizes. Results are recorded when one team of agents is entirely defeated. After multiple experiments, victory rates and other relevant metrics are calculated. To promote fair confrontation, we provide each team with exactly the same initial supplies. As shown in Fig.~\ref{fig:pvp}, this setup clearly demonstrate and quantify the differences of these two methodologies under competitive conditions. The parallelized framework demonstrates a significant advantage over the serialized framework in this dynamic adversarial scenario. Our analysis indicates that this advantage is primarily attributed to our interruption mechanism, which enables agents to dynamically adjust their strategies and respond promptly to changes in the environment (\textit{e.g.}, seamlessly switch attack targets, prioritize health restoration, more efficient communication, etc.).

\subsection{Ablation Study}
\label{sec:ablation}
\subsubsection{Recursive Task Decomposition}

We first evaluate the effectiveness of our skill library's recursive task decomposition mechanism by assessing a single agent's performance on resource collection tasks in Minecraft. This evaluation aimed to verify the comprehensive skill library's capability in automating complex workflows, thereby validating its efficiency and reliability in handling fundamental Minecraft tasks. As shown in Table.~\ref{tab:skill_performance_comparison}, when the recursive task decomposition mechanism is ablated, the system can only complete short-term tasks that require fewer steps and shows a reduction in efficiency. In contrast, utilizing our comprehensive skill library with the recursive task decomposition mechanism enables efficient completion of all tasks.

\begin{table}[htbp]
\centering
\caption{Average completion time with standard deviation and success rate (SR) comparison with and without the recursive task decomposition mechanism (RTDM). Tasks from top to bottom: (1) crafting table, (2) wooden tool, (3) stone tool, (4) iron tool, (5) diamond.}
\label{tab:skill_performance_comparison}
\begin{tabular}{lcccc}
\toprule
\textbf{Task} & \multicolumn{2}{c}{\textbf{with RTDM}} & \multicolumn{2}{c}{\textbf{w/o RTDM}} \\
\cmidrule(lr){2-3} \cmidrule(lr){4-5}
& Time~(min) & SR & Time~(min) & SR \\
\midrule
\includegraphics[width=0.4cm]{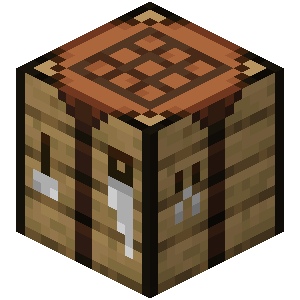} & $0.3 \pm 0.2$ & $100\%$ & $2.8 \pm 2.5$ & $100\%$ \\
\includegraphics[width=0.4cm]{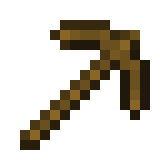} & $0.6 \pm 0.3$ & $100\%$ & $4.6 \pm 2.5$ & $80\%$ \\
\includegraphics[width=0.4cm]{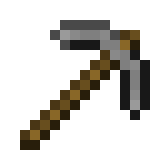} & $1.4 \pm 0.5$ & $100\%$ & $N/A$ & $0\%$ \\
\includegraphics[width=0.4cm]{fig/MC/IronPickaxe.png} & $4.7 \pm 1.3$ & $100\%$ & $N/A$ & $0\%$ \\
\includegraphics[width=0.4cm]{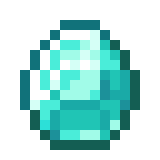} & $6.2 \pm 1.6$ & $100\%$ & $N/A$ & $0\%$ \\
\bottomrule
\end{tabular}
\end{table}

In the more complex resource collection task we propose in Section ~\ref{sec:resource}, the MAS will not be able to make any meaningful progress without the recursive task decomposition mechanism, suggesting that invoking an LLM to independently plan the task dependencies embedded within a DAG would not be as efficient as utilizing a fixed mechanism.

\subsubsection{Parallelized Framework}
To validate the necessity of our method, we perform ablation studies in boss combat tasks by disabling different components in our framework:

\begin{itemize}[leftmargin=*, itemsep=2pt]
\item \textbf{w/o Parallelized Planning-Acting Framework (PPA)}: Replaced our parallelized framework with traditional serialized execution, disabling the interruption mechanism.

\item \textbf{w/o Centralized Memory System (CMS)}: Disabled real-time team observation polling, chat logs, action history and global progress information, thereby restricting agents to rely solely on their individual observations.
\end{itemize}

\begin{figure}[!t]
\centering
\includegraphics[width=0.5\textwidth]{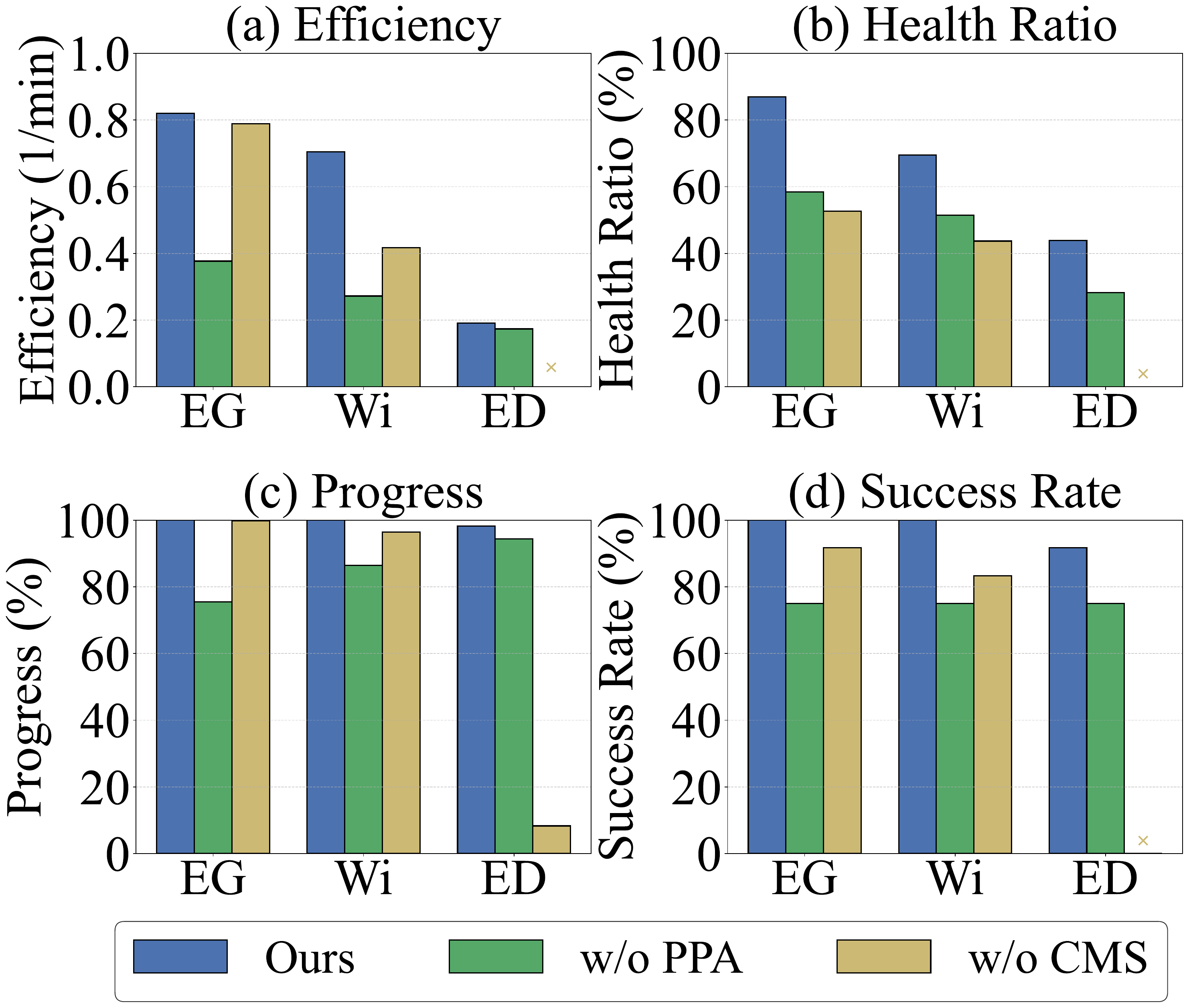}
\caption{Ablation study in boss combat tasks across 12 trials in three scenarios: Elder Guardian (EG), Wither (Wi), and Ender Dragon (ED), where efficiency is defined as the inverse of completion time in minutes.}
\label{fig:ablation}
\end{figure}

% These ablation configurations systematically evaluate the contribution of each architectural component to overall system performance, providing insights into their relative importance in complex multi-agent coordination tasks. 
Experiment results shown in Fig.~\ref{fig:ablation} highlight the critical roles of the parallelized planning-acting framework and the centralized memory system. Besides these ablation studies, our method also performs exceptionally well in direct comparative experiments. Please refer to Appendix for details.

\begin{table*}[htbp]
    \centering
    \caption{Performance of boss combat across three task scenarios using different observation modalities, with a fixed team size of 10 agents. Reporting mean values and standard deviations of multiple evaluation metrics over 12 trials.}
    \label{tab:boss_combat_modality}
    \begin{tabular*}{\textwidth}{@{\extracolsep{\fill}}cccccc}
        \toprule
        \textbf{Task Scenario} & \textbf{Observation Modality} & \textbf{Time(min)} & \textbf{Health Ratio} & \textbf{Progress} & \textbf{Success Rate} \\ 
        \midrule

        \multirow{2}{*}{Elder Guardian}
        & Text & $1.2 \pm 0.3$ & $86.9 \pm 8.4\%$ & $100.0 \pm 0.0\%$ & $100.0\%$ \\
        & Visual & $1.9 \pm 0.7$ & $85.3 \pm 6.1\%$ & $100.0 \pm 0.0\%$ & $100.0\%$ \\
        \midrule

        \multirow{2}{*}{Wither}
        & Text & $1.4 \pm 0.3$ & $69.5 \pm 19.0\%$ & $100.0 \pm 0.0\%$ & $100.0\%$ \\
        & Visual & $2.3 \pm 0.5$ & $39.1 \pm 12.4\%$ & $82.8 \pm 7.2\%$ & $66.7\%$ \\
        \midrule

        \multirow{2}{*}{Ender Dragon}
        & Text & $5.2 \pm 1.5$ & $43.9 \pm 24.4\%$ & $98.2 \pm 5.9\%$ & $91.7\%$ \\
        & Visual & $6.1 \pm 1.1$ & $21.8 \pm 10.3\%$ & $58.3 \pm 13.5\%$ & $50.0\%$ \\
        \bottomrule
    \end{tabular*}
\end{table*}

\subsection{Robustness Analysis}

\subsubsection{Robust to Modality}
We evaluated the framework's robustness to different modalities by replacing the text-based observation with a multi-modal approach in boss combat tasks. While the inclusion of visual language models (VLMs) introduced increased response latency and slightly reduced observation accuracy, the agents maintained reasonable performance levels. As summarized in Table~\ref{tab:boss_combat_modality}, our framework achieves strong performance across all three challenging task scenarios using either modalities.

\subsubsection{Robust to Scale}
To validate the effectiveness of our approach under large-scale scenarios, we conducted additional experiments involving teams of 5-50 agents. These experiments focused on measuring the impact of agent quantity (where the number of agents equaled the number of latest chat entries read per LLM inference) on token counts and inference time.

\begin{figure}[!t]
    \centering
    \includegraphics[width=0.5\textwidth]{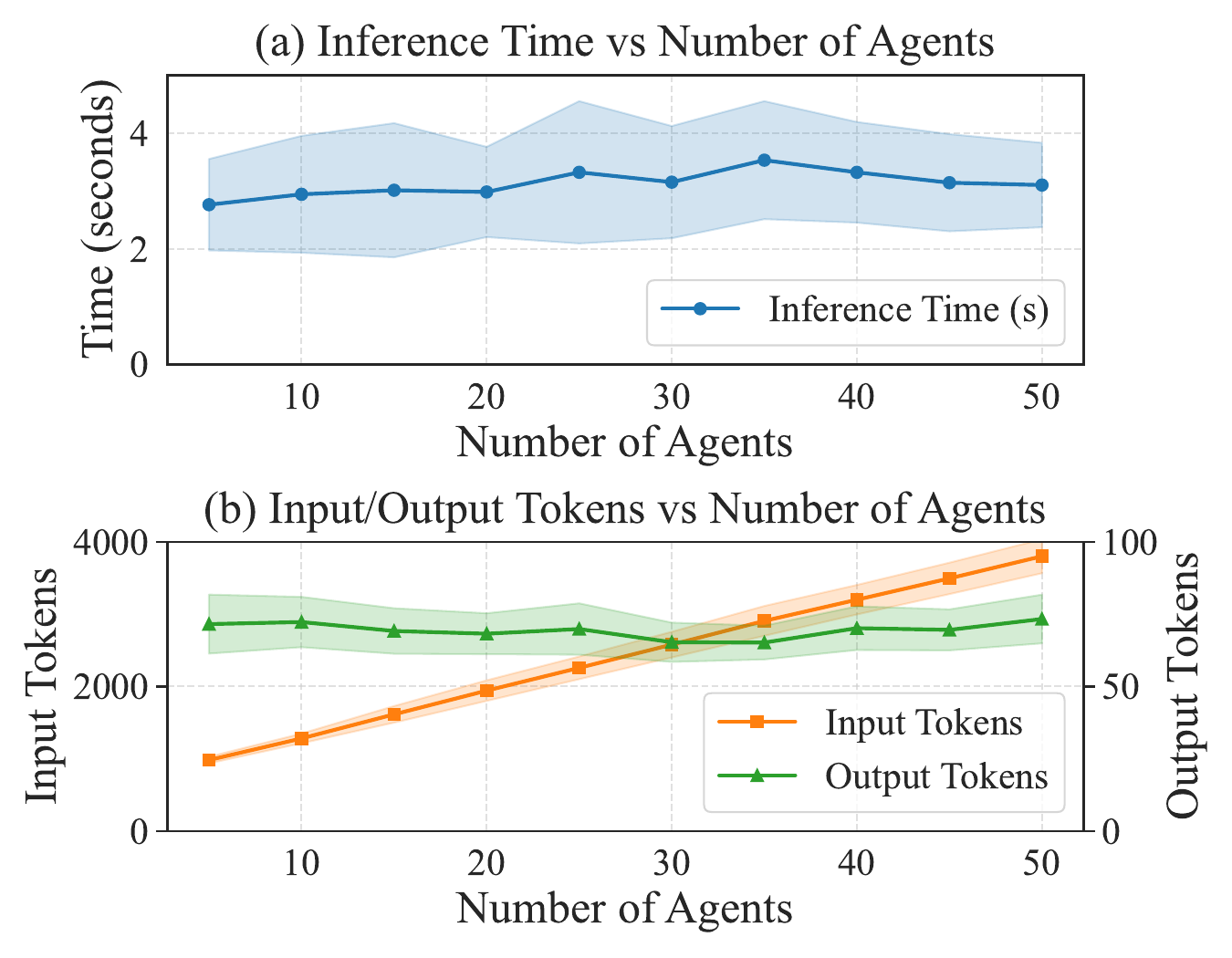}
    \caption{Inference time and input/output tokens per single-agent planning step of MAS at different scales.}
    \label{fig:scale}
\end{figure}

In Fig.~\ref{fig:scale}, it can be observed that the inference time of the LLM tends to stabilize instead of increasing continuously as the number of agents increases. Additionally, the total token cost shows an approximately linear growth with the number of agents, which is both acceptable and manageable. Since each agent is equipped with parallelized planning threads and acting threads independently, the actions of each agent remain unaffected by the total number of agents in the system. Further analysis can be found in the Appendix.

% This may be attributed to the fact that we utilize the Qwen-Plus API, which likely incorporates advanced optimization techniques such as vLLM~\cite{vllm} acceleration, enabling the system to efficiently handle increased tokens without significantly impacting inference latency, thereby maintaining stable performance even as the input size grows.
The input tokens increases linearly with the number of agents, as more chat entries are processed during each planning step. However, the output tokens remain stable due to the constraints imposed by the prompt format, which limits the length of generated responses. Overall, the system's total latency does not increase significantly with the number of agents, and the total cost (tokens) grows approximately linearly with the number of agents, which is acceptable and controllable.

\section{Related Works}

\textbf{LLM-based Minecraft Agents.} The development of LLM-based AI agents in Minecraft has evolved through several key approaches: Voyager~\cite{voyager} established the first LLM-based agent with automatic skill discovery using GPT-4~\cite{gpt4}, built on the open-source library Mineflayer~\cite{Mineflayer}. Subsequent studies enhanced agents via specialized memory mechanisms~\cite{mrsteve, optimus}, specialized LLM fine-tuning~\cite{llamarider, steve, odyssey}, task decomposition and causal graph learning~\cite{plan4mc, gitm, adam}, and combination with reinforcement learning~\cite{larm, auto-mcreward}. Additionally, multi-modal information perception and processing were explored~\cite{steve-eye, rocket-1}, along with other novel techniques~\cite{optimus3, omnijarvis, minedreamer}. The development of benchmarks for general capabilities progressed with MineDojo~\cite{minedojo} and MCU~\cite{mcu}, while specific agent capabilities were assessed through additional benchmarks~\cite{mp5, villager, mineanybuild}.

\textbf{LLM-based Multi-Agent Systems.} Recent research has focused on several core areas~\cite{ma-survey1,ma-survey2}: infrastructure frameworks for efficient agent coordination~\cite{metagpt,agentverse,coela,proagent,supernet}, which introduce novel paradigms for task management and team collaboration; efforts aimed at improving MAS collaboration efficiency through optimizations at communication or routing~\cite{masrouter,cut-the-crap,chen2024optima}; benchmark development to evaluate multi-agent performance in dynamic environments~\cite{llmarena, villager}, which has created robust testing environments to assess the generalization and efficiency of LLM-powered agents; large-scale social simulations~\cite{sid, gen-agents, oasis}, which explore how multi-agent systems model complex societal behaviors; and domain-specific applications~\cite{jubensha,marg,multirobot} demonstrating the effectiveness of LLM-based agents in specific scenario simulation. In contrast to prior MAS research that primarily focuses on coordination paradigms or communication strategies, our work targets continuous real-time interaction in non-paused dynamic environments.

\section{Conclusion}

We propose a novel parallelized planning-acting multi-agent framework that significantly enhances the responsiveness and adaptability of LLM-based MAS in dynamic environments like Minecraft. Our framework's dual-thread architecture with interruptible execution mechanism enables real-time interaction and continuous adaptation, overcoming the limitations of traditional serialized execution paradigms. The comprehensive skill library and recursive task decomposition mechanism further improve efficiency and coordination as an engineering contribution. Experiments on our challenging benchmark tasks validate the effectiveness of our framework in diverse Minecraft scenarios.

\begin{acks}
This work is supported in part by the Hangzhou Joint Funds of the Zhejiang Provincial Natural Science Foundation of China under Grant No. LHZSD24F020001, in part by the Zhejiang Province High-Level Talents Special Support Program ``Leading Talent of Technological Innovation of Ten-Thousands Talents Program'' under Grant No. 2022R52046, in part by the Fundamental Research Funds for the Central Universities under Grant No. 2021FZZX001-23, and in part by the advanced computing resources provided by the Supercomputing Center of Hangzhou City University.
\end{acks}

\ifarxiv

\else
\bibliographystyle{ACM-Reference-Format} 
\bibliography{sample}
\fi

%%%%%%%%%%%%%%%%%%%%%%%%%%%%%%%%%%%%%%%%%%%%%%%%%%%%%%%%%%%%%%%%%%%%%%%%

\ifarxiv

\lstset{% 
	basicstyle={\footnotesize\ttfamily},
	numbers=left,numberstyle=\footnotesize,xleftmargin=2em,
	aboveskip=0pt,belowskip=0pt,
	showstringspaces=false,tabsize=2,breaklines=true}
\floatstyle{ruled}
\newfloat{listing}{tb}{lst}{}
\floatname{listing}{Listing}

% 定义浅色背景和深色标题背景
\definecolor{lightbluecolor}{RGB}{240, 248, 255} % 浅蓝色背景
\definecolor{darkbluecolor}{RGB}{0, 51, 102}  % 深蓝色标题背景

% 定义浅绿色和深绿色
\definecolor{lightgreencolor}{RGB}{235, 245, 235} % 浅绿色背景
\definecolor{darkgreencolor}{RGB}{0, 102, 51}     % 深绿色标题背景

\newtcolorbox{bluebox}[1][]{
    colback=lightbluecolor,         % 背景颜色
    colframe=darkbluecolor,      % 边框颜色
    colbacktitle=darkbluecolor,  % 标题背景颜色
    coltitle=white,               % 标题文字颜色
    fonttitle=\bfseries,          % 标题字体加粗
    breakable,                    % 允许盒子内容跨页
    enhanced,                     % 启用增强功能
    width=\linewidth,             % 设置宽度为文本宽度
    boxrule=0.5mm,                % 边框厚度
    arc=2mm,                      % 圆角半径
    auto outer arc,               % 自动调整外圆角
    #1                             % 允许传递额外选项
}

\newtcolorbox{greenbox}[1][]{
    colback=lightgreencolor,      % 背景颜色
    colframe=darkgreencolor,      % 边框颜色
    colbacktitle=darkgreencolor,  % 标题背景颜色
    coltitle=white,               % 标题文字颜色
    fonttitle=\bfseries,          % 标题字体加粗
    breakable,                    % 允许盒子内容跨页
    enhanced,                     % 启用增强功能
    width=\linewidth,             % 设置宽度为文本宽度
    boxrule=0.5mm,                % 边框厚度
    arc=2mm,                      % 圆角半径
    auto outer arc,               % 自动调整外圆角
    left=2mm, right=2mm, top=2mm, bottom=2mm, % 内边距
    #1                             % 允许传递额外选项
}

% 配置 listings 宏包
\lstset{
    basicstyle=\normalfont\normalsize, % 设置字体样式，使用正常大小
    backgroundcolor=\color{lightbluecolor},   % 背景颜色设为与 bluecolorbox
    frame=none,                      % 移除 lstlisting 的边框
    breaklines=true,                 % 自动换行
    showstringspaces=false,          % 不显示空格
    numbers=none,                    % 不显示行号
    tabsize=2,                       % 设置制表符宽度
    xleftmargin=0pt,                 % 减少左侧边距
    xrightmargin=0pt,                % 减少右侧边距
    aboveskip=0pt,                   % 减少上方间距
    belowskip=0pt                    % 减少下方间距
}

\newpage
\appendix
\section{Centralized Memory System}
\label{appendix:memory}

The memory of the multi-agent system is stored in JSON files, and agents read and modify their contents by reading and writing to these files. 

The observations and states of the agent are updated in a polling manner per second and saved in the following form, constantly overwritten.

\begin{bluebox}[title=Observation Records]
\begin{lstlisting}
{
  "time": "day",
  "health": 12.8,
  "hunger": 17,
  "position": "(10, 65, -8)",
  "equipment": [
    null,
    null,
    null,
    null,
    "bow",
    null
  ],
  "inventory": {
    "arrow": 187,
    "golden_apple": 4,
    "diamond_pickaxe": 1,
    "diamond_sword": 1,
    "diamond_helmet": 1,
    "diamond_chestplate": 1,
    "diamond_leggings": 1,
    "diamond_boots": 1,
    "bow": 1
  },
  "nearbyBlocks": [
    "end_stone",
    "bedrock"
  ],
  "nearbyEntities": "enderman, ender_dragon",
  "nearbyPlayers": [
    "Steve",
    "Brine",
    "Jeb",
    "Notch"
  ]
}
\end{lstlisting}
\end{bluebox}

The team chat logs are constantly updated by active communication and continue to grow in real time, but only the most recent information is taken as prompts.

\begin{bluebox}[title=Chat Logs (Active Communication)]
\begin{lstlisting}
{
  "Day 1 - Morning": [
    "SystemInfo: Number of End Crystals Remaining: 10",
    "SystemInfo: Boss Health: 200",
    "SystemInfo: Boss Health: 200",
    "Notch: Let's prioritize taking down the end_crystals to prevent the Ender Dragon from healing. I will start attacking the nearest end_crystal with my bow. Everyone, cover me and be ready to assist if needed.",
    "Steve: I'm ready with my bow and arrows. Let's focus on taking down the end_crystals first to prevent the Ender Dragon from healing. I'll start targeting the nearest end_crystal. Stay alert for any incoming endermen but avoid engaging them unless necessary.",
    "SystemInfo: Boss Health: 200",
    "Brine: Brine: Let's prioritize destroying the end_crystals to prevent the Ender Dragon from healing. Notch and I will start targeting the nearest crystal. Steve, Alex, and Jeb, please spread out and assist in taking down other crystals. Stay alert for any Endermen but avoid engaging them unless necessary. Report if you need help or spot any crystals.",
    "Jeb: I'm ready with my bow and arrows. Let's prioritize the end_crystal on the obsidian pillars to prevent the Ender Dragon from healing. I'll start targeting the nearest crystal. Stay alert for any incoming endermen but avoid engaging unless necessary."
    ...
  ]
}
\end{lstlisting}
\end{bluebox}

The team chat logs are also continuously updated in real time by passive communication, but only the latest observations of each agent are provided to the LLM as prompts.

\begin{bluebox}[title = Chat Logs (Passive Communication)]
\begin{lstlisting}
{
  "obs_entries": [
    "Notch: Team, I'm fully equipped with diamond armor, a bow, and 200 arrows. I see an end_crystal and an ender_dragon nearby, along with some endermen. Brine, Jeb, Steve, and Alex are also close. Let's focus on taking down the end_crystal first to prevent the Ender Dragon from healing. Everyone, prepare your bows and stay alert for any incoming endermen. Ready when you are!",
    ...
    "Brine: Team, I'm currently well-equipped with a bow, 200 arrows, and diamond armor. The ender_dragon is nearby but hasn't attacked yet. I see Steve, Notch, Alex, and Jeb around. Let's focus on taking out the end_crystals first. I suggest we spread out to cover more ground and target the crystals from a distance. Stay alert for any Endermen, but avoid engaging them unless necessary. Report if you spot any crystals or need assistance.",
    ...
    "Jeb: All end_crystals have been destroyed. Focusing on the Ender Dragon now. Everyone, keep your distance and use bows to attack from a safe range. Stay alert for any incoming endermen but prioritize the Ender Dragon. Let's finish this!",
    ...
  ]
}
\end{lstlisting}
\end{bluebox}

The team action logs record the historical actions and reasoning processes of each agent, but only an agent's own current action is used as LLM prompts.

\begin{bluebox}[title = Action History]
\begin{lstlisting}
{
    "Notch": [
        "```json\n{\n    \"skill\": \"chatMessage(bot, \\\"Let's prioritize taking down the end_crystals to prevent the Ender Dragon from healing. I will start attacking the nearest end_crystal with my bow. Everyone, cover me and be ready to assist if needed.\\\", 'A')\",\n    \"interrupt\": false,\n    \"reason\": \"Communicate the plan to the team before starting the attack on the end_crystal.\"\n}\n```",
        "```json\n{\n    \"skill\": \"combatWithEntity(bot, 'end_crystal', 'bow', true)\",\n    \"interrupt\": false,\n    \"reason\": \"Start attacking the nearest end_crystal with my bow to prevent the Ender Dragon from healing. The team has been informed and is prepared to assist if needed.\"\n}\n```",
        "```json\n{\n    \"skill\": \"combatWithEntity(bot, 'end_crystal', 'bow', true)\",\n    \"interrupt\": false,\n    \"reason\": \"I am currently attacking the nearest end_crystal with my bow to prevent the Ender Dragon from healing. My health is full, and there are no immediate threats requiring my attention. Continuing this action is crucial for our strategy.\"\n}\n```",
        ...
    ],
    "Steve": ...
}
\end{lstlisting}
\end{bluebox}

In addition to this, agents are also provided with identity information such as their name, team affiliation, and teammates' names, to simulate a human-like multi-agent collaborative scenario.

\section{Comprehensive Skill Library}
\label{appendix:skill}
Our proposed comprehensive skill library for Minecraft is implemented based on Mineflayer. To simulate the process of solving prerequisite steps in the actual game to eventually obtain specific items, this skill library has introduced a recursive task decomposition mechanism. From the implementation level, all interfaces can recursively invoke each other or themselves and pass parameters.

\subsection{Minecraft Skill APIs}
\label{appendix:skill-api}
The \textbf{Basic Skill Interfaces} provide the foundational functions required to perform essential actions in the game, such as collecting items, crafting, mining, smelting, and combat. These functions are designed to be modular and reusable, allowing for seamless integration into more complex workflows.

\begin{greenbox}[title = Basic Skill Interfaces]
\begin{itemize}
    \item \texttt{obtainItem(bot, count, type)}: Automates the collection of items.
    \item \texttt{mineItem(bot, count, type, explore\_direction, explore\_time)}: Mines using specific tools. If the exploration parameters are not specified, the exploration direction is chosen randomly. If the item type is a subterranean mineral like diamond, the exploration direction is set to down (0, -1, 0). The exploration time limit defaults to five minutes (6000 ticks in Minecraft).
    \item \texttt{craftItem(bot, count, type, need\_crafting\_table)}: Crafts items. If no parameter is specified, the crafting is done using a crafting table by default.
    \item \texttt{smeltItem(bot, count, type, fuel)}: Smelts or cooks items in a furnace. If no parameters are provided, coal is used as the default fuel.
    \item \texttt{collectItem(bot, count, type, function)}: Collects items by killing animals or using other special methods.
    \item \texttt{chatMessage(bot, message, team\_name)}: Sends a chat message to the team.
    \item \texttt{getItemFromChest(bot, chest\_position, items\_to\_get)}: Retrieves specific items from a chest at a given location.
    \item \texttt{depositItemIntoChest(bot, chest\_position, items\_to\_deposit)}: Deposits specific items into a chest at a given location.
    \item \texttt{combatWithEntity(bot, mob\_name, weapon, loop)}: Automatically equips the highest quality armor and weapon from the inventory to fight with an entity (\textit{e.g.}, animals or hostile mobs). If no parameters are specified, the default weapon is a sword. The \texttt{loop} parameter determines whether to continue fighting the same type of entity.
    \item \texttt{combatWithPlayer(bot, player\_name, weapon)}: Automatically equips the highest quality armor and weapon from the inventory to fight with a specific player. If no parameters are provided, the default weapon is a sword.
    \item \texttt{initialInventory(bot, item\_dict)}: Initializes the player's inventory with specified items.
    \item \texttt{equipBestToolOrArmor(bot, type)}: Automatically equips the highest quality tool, weapon, or armor of the specified type.
    \item \texttt{listenChat(bot, player\_name)}: Continuously listens to a player’s chat messages.
\end{itemize}
\end{greenbox}

The \textbf{Recursive-related Interfaces} are designed to handle the dependencies between tasks. These functions determine the prerequisites for obtaining specific items, such as tools, materials, or entities, and ensure that all necessary steps are completed before proceeding with the main task. 

All task dependencies are modeled as a Directed Acyclic Graph (DAG), where nodes represent tasks and edges represent prerequisite relationships. We have cataloged thousands of these intrinsic dependencies within Minecraft to support this recursive decomposition mechanism. This not only provides a methodological framework but also constitutes a significant engineering contribution by facilitating the automation of complex task sequences.

\begin{greenbox}[title = Recursive-related Interfaces]
\begin{itemize}[leftmargin=*, itemsep=2pt]
    \item \texttt{preTool(item)}: Retrieves the minimum prerequisite tool required for a specific item.
    \item \texttt{preItem(item)}: Retrieves the prerequisite items required to craft a specific item and indicates whether a crafting table is necessary.
    \item \texttt{preSmelt(item)}: Retrieves the prerequisite items required to smelt a specific item.
    \item \texttt{preCollect(item)}: Retrieves the prerequisite entities required for a specific item.
    \item \texttt{getFunc(item)}: Retrieves the method used to collect a specific item.
\end{itemize}
\end{greenbox}

\subsection{Implementation Details}
\label{appendix:skill-imp}
Algorithm~\ref{alg:obtain} demonstrates the implementation of the \texttt{obtainItem} function, which determines the collection method for a specific item and delegates the task to the corresponding sub-function (\textit{e.g.}, mining, crafting, smelting). This function serves as the entry point for recursive task decomposition.

\begin{algorithm}[H]
\caption{obtainItem}
\label{alg:obtain}
\begin{algorithmic}[1]
\Function{obtainItem}{$bot, cnt, type$}
    \State $func \gets \Call{getCollectionMethod}{type}$
    \If{$func = \text{``craft''}$}
        \State \Return $\Call{craftItem}{bot, cnt, type}$
    \ElsIf{$func = \text{``mine''}$}
        \State \Return $\Call{mineItem}{bot, cnt, type}$
    \ElsIf{$func = \text{``smelt''}$}
        \State \Return $\Call{smeltItem}{bot, cnt, type}$
    \Else
        \State \Return $\text{false}$
    \EndIf
\EndFunction
\end{algorithmic}
\end{algorithm}

The \texttt{mineItem} function shown in Algorithm~\ref{alg:mine} handles the process of mining blocks or ores using appropriate tools. If the required tool is not in the inventory, it recursively calls \texttt{obtainItem} to craft or gather the tool before proceeding with mining. This ensures that the bot is always equipped with the necessary tools for the task.

\begin{algorithm}[H]
\caption{mineItem}
\label{alg:mine}
\begin{algorithmic}[1]
\Function{mineItem}{$bot, cnt, type, dir$}
    \State $tool \gets \Call{preTool}{type}$
    \If{$\Call{invCnt}{tool} = 0$}
        \State $\Call{obtainItem}{bot, 1, tool}$
    \EndIf
    \State $\Call{equip}{bot, tool}$
    \If{$dir = \text{null}$}
        \State $dir \gets \Call{randomDirection}{}$
    \EndIf
    \State $\Call{startMining}{bot, type, dir, time}$
    \State \Return $\Call{verifyResult}{cnt, type}$
\EndFunction
\end{algorithmic}
\end{algorithm}

The \texttt{craftItem} function shown in Algorithm~\ref{alg:craft} retrieves the recipe requirements for the specified item. If the required materials are insufficient, it recursively calls \texttt{obtainItem} to gather them before proceeding with crafting.

\begin{algorithm}[H]
\caption{craftItem}
\label{alg:craft}
\begin{algorithmic}[1]
\Function{craftItem}{$bot, cnt, type$}
    \State $reqs \gets \Call{getRecipeReq}{type}$
    \For{$(r\_cnt, r\_type)$ in $reqs$}
        \If{$\Call{invCnt}{r\_type} < r\_cnt$}
            \State $n \gets r\_cnt - \Call{invCnt}{r\_type}$
            \State $\Call{obtainItem}{bot, n, r\_type}$
        \EndIf
    \EndFor
    \State $\Call{executeCrafting}{type, cnt}$
    \State \Return $\Call{verifyResult}{cnt, type}$
\EndFunction
\end{algorithmic}
\end{algorithm}

The \texttt{smeltItem} function shown in Algorithm~\ref{alg:smelt} checks if input items for smelting are available. If not, it recursively calls \texttt{obtainItem} to gather necessary materials before proceeding with smelting.

\begin{algorithm}[H]
\caption{smeltItem}
\label{alg:smelt}
\begin{algorithmic}[1]
\Function{smeltItem}{$bot, cnt, type$}
    \State $input \gets \Call{getSmeltingInput}{type}$
    \If{$\Call{invCnt}{input} < cnt$}  
        \State $n \gets cnt - \Call{invCnt}{input}$
        \State $\Call{obtainItem}{bot, n, input}$
    \EndIf
    \State $\Call{prepareFurnaceAndFuel}{}$
    \State $\Call{executeSmelting}{type, cnt}$
    \State \Return $\Call{verifyResult}{cnt, type}$
\EndFunction
\end{algorithmic}
\end{algorithm}

\section{Experiment Details}
\label{appendix:exp}
All experiments were conducted using the Qwen-Plus and Qwen-VL-plus model API provided by Alibaba Cloud. In the multiplayer local area network (LAN) server of Minecraft Java Edition version 1.19.4, agents connected to the game via different ports as an independent player and interacted with the environment and other agent players.

\subsection{Resource Collection Task}
\label{appendix:resource}
Our framework supports tasks formulated as collecting any quantity of various types of items, represented by a requirement dictionary. This design ensures the richness and flexibility of supported tasks, enabling multi-agent collaboration to automatically collect resources based on given requirements. Such a setup makes our benchmark both general and scalable. However, due to inevitable time and API costs, we selected a set of representative resource collection tasks for our experiments. These tasks not only demonstrate the capabilities of our framework but also highlight common challenges faced by players in Minecraft, such as gathering essential survival tools (the \textit{Iron Tool Set} task) or preparing for combat with powerful boss monsters (the \textit{Diamond Armor} task). The following are the representative resource collection tasks we experimented with:

\begin{greenbox}[title = Task Definitions]
\begin{itemize}
    \item \textbf{Iron Tool Set:} \{ 'iron\_pickaxe': 1, 'iron\_shovel': 1, 'iron\_hoe': 1, 'iron\_axe': 1 \} — A set of commonly used iron tools in Minecraft.
    \item \textbf{Diamond Armor:} \{ 'diamond\_helmet': 1, 'diamond\_chestplate': 1, 'diamond\_leggings': 1, 'diamond\_boots': 1 \} — A full set of diamond armor in Minecraft.
    \item \textbf{Redstone Devices:} \{ 'repeater': 1, 'piston': 1, 'dropper': 1 \} — Common redstone components and devices in Minecraft.
    \item \textbf{Navigation Kit:} \{ 'compass': 1, 'clock': 1, 'map': 1 \} — A set of tools commonly used for navigation in Minecraft.
    \item \textbf{Transport System:} \{ 'minecart': 1, 'rail': 16, 'powered\_rail': 6 \} — Minecarts and tracks used to complete a transportation system.
    \item \textbf{Food Supplies:} \{ 'beef': 1, 'chicken': 1, 'porkchop': 1 \} — Common animal meat food items in Minecraft.
    \item \textbf{Building Materials:} \{ 'stone\_bricks': 4, 'glass': 4, 'iron\_door': 1 \} — Building materials commonly used for constructing doors, windows, and walls.
    \item \textbf{Storage System:} \{ 'hopper': 1, 'chest': 1, 'barrel': 1 \} — Tools commonly used for storing items in Minecraft.
\end{itemize}
\end{greenbox}

Each round of the experiment is conducted in a world generated with a random seed to ensure the generalizability of the results.

\subsection{Boss Combat Task}
\label{appendix:combat}
In Minecraft, there are three primary world dimensions: Overworld, Nether, and End. Each dimension hosts powerful boss monsters whose defeat is considered the pinnacle challenge for players of Minecraft. Based on this, we have predefined three combat tasks:

\begin{itemize}[leftmargin=*, itemsep=2pt]
    \item \textbf{Elder Guardian (Overworld):} This task involves defeating the formidable boss \textit{ Elder Guardian} within complex underwater terrain of an ocean monument. Agents must also contend with other surrounding monsters, adding complexity that requires the MAS to develop effective combat strategies.
    
    \item \textbf{Wither (Nether):} This task involves battling the powerful boss \textit{Wither} in the Nether. Agents face additional challenges from smaller enemies such as piglins and ghasts, amidst a landscape filled with lava and other hazards. When the Wither's health drops below half, it enters a berserk state where it becomes immune to ranged attacks, necessitating the MAS to dynamically adjust combat strategies based on real-time observations.

    \item \textbf{Ender Dragon (End):} Defeating the \textit{Ender Dragon} in the End is considered the ultimate challenge in Minecraft. Multiple agents need to strategically cooperate by first destroying the end crystals located atop obsidian pillars to disable the dragon’s healing mechanism before engaging in battle with the dragon and its surrounding endermen, requiring the MAS to adjust tactics adaptively.
\end{itemize}

Prior to each combat, agents were equipped with standardized combat resources, including a full set of diamond armor (validated as efficiently collectible through our resource collection experiments), bow and arrows, and some consumables. In all experiments, the standardized combat resources remained consistent. A single LLM call initialized resource allocation, distributing equipment to each agent through a dictionary-based assignment system. Agents were then teleported to designated combat locations, where boss monsters and supporting entities (\textit{e.g.}, Guardians near the Elder Guardian, End Crystals for Ender Dragon healing) were spawned to ensure scenario complexity. During combat, real-time progress information (boss health, remaining End Crystals, etc.) was communicated through the chat system to support strategic planning. Task completion time was measured excluding initialization phases.

Evaluation metrics are defined as follows:
\begin{equation}
    \text{health ratio} = \frac{1}{n}\sum_{i=1}^n \frac{h_i}{H_{\text{max}}} \times 100\% 
\end{equation}
where $ h_i $ is the remaining health of agent $ i $, and $ H_{\text{max}} $ is the maximum health of the agent.

\begin{equation}
    \text{progress} = \frac{H_{\text{boss}} - h_{\text{boss}}}{H_{\text{boss}}} \times 100\%
\end{equation}
where $ h_{\text{boss}} $ is the remaining health of the boss monster, and $ H_{\text{boss}} $ is the maximum health of the boss monster.

\subsection{Adversarial PVP Task}

In addition to ablation studies, we design a task scenario where two frameworks can be directly compared, namely Adversarial PVP Task. Similar to the Boss Combat Task, in this task, agents were provided with combat resources. Then, the two teams of agents engaged in battles with each other. Results were recorded when one team of agents was entirely defeated. After multiple experiments, victory rates and other relevant metrics were calculated. In order to promote fair confrontation, we give each team exactly the same initial supplies. Here is an example of initial inventory given to an agent : \{
    'wooden\_sword': 1,
    'diamond\_helmet': 1,
    'diamond\_chestplate': 1,
    'diamond\_leggings': 1,
    'diamond\_boots': 1,
    'bow': 1,
    'arrow': 200,
    'golden\_apple': 20
\}

\subsection{Ablation Study}

Our ablation studies respectively focused on the comprehensive skill library and its recursive task decomposition, the centralized memory system and the parallelized planning-acting mechanism, which are three core contributions proposed in our paper. Although it is challenging to conduct fair comparisons between different approaches and frameworks in agent-based research due to varying architectures and objectives, each ablation here can be regarded as a methodological baseline. For example, ablating the parallelized planning-acting mechanism is actually taking the traditional serialized planning-acting architecture as the baseline.

\subsection{Scale Robustness Analysis}
\label{appendix:roboust}

Our method theoretically supports multi-agent systems of any scale. Regardless of the scale, 
each agent is independently equipped with parallel planning threads and acting threads and interacts with the environment in real time. As for the acting thread of a single agent, its actions are completely unaffected by the number of agents in the entire system. However, for the planning thread, under our settings, the length of the chat log window input for each LLM call is equal to the number of system agents, which means that as the system scale increases, the number of input tokens for LLM inference continues to increase. So we conducted an experiment on this. By adjusting the window length of the chat log, we sampled different numbers of chat entries for LLM planning and counted their inference duration and the number of input and output tokens. 

As shown in the main paper, experiment results show that the LLM inference time tends to stabilize rather than grow continuously with increasing agents and the total token cost grows approximately linearly with the number of agents. This may be attributed to the fact that we utilize the Qwen-Plus API, which likely incorporates advanced optimization techniques such as vLLM acceleration, enabling the system to efficiently handle increased tokens without significantly impacting inference latency, thereby maintaining stable performance even as the input size grows. The input tokens increases linearly with the number of agents, as more chat entries are processed during each planning step. However, the output tokens remain stable due to the constraints imposed by the prompt format, which limits the length of generated responses. Overall, our method demonstrates robustness to the scale of multi-agent systems.

\section{Human-Agent Interaction}
\label{appendix:sec:human}
Our framework also supports flexible human-agent interaction, allowing human players to either actively participate in task execution alongside agents, facilitating dynamic cooperation within the game environment, or to take on a guiding role by providing instructions to agents for task division, coordination, and strategic decision-making. This section presents examples of human-agent interactions in Minecraft, highlighting how AI agents and human players collaborate in various tasks.

\subsubsection{Scenario 1: Human-Agent Collaboration}

\begin{bluebox}[title = Agent Steve's Environment Information]
\begin{lstlisting}
{
  "time": "day",
  "health": 20,
  "hunger": 20,
  "position": "(-223, 60, -159)",
  "equipment": [
    null,
    null,
    null,
    null,
    "crafting_table",
    null
  ],
  "inventory": {
    "crafting_table": 1
  },
  "nearbyBlocks": [
    "dirt",
    "grass_block",
    "oak_log",
    "oak_leaves"
  ],
  "nearbyEntities": "pig",
  "nearbyPlayers": ["Alex", "Notch"]
}
\end{lstlisting}
\end{bluebox}

\begin{greenbox}[title = Human Player's Chat Message]
'Hey Steve, I've got some beef, can you hunt a pig for obtaining porkchops?'
\end{greenbox}

\begin{bluebox}[title = Agent Steve's Planning Action]
\begin{lstlisting}
{
    "skill": "obtainItem(bot, 10, \'porkchop\')",
    "interrupt": false,
    "reason": "Following the human player's request to hunt a pig for obtaining porkchops. Since there are pigs nearby, I will proceed with hunting to gather porkchops."
}
\end{lstlisting}
\end{bluebox}

\subsubsection*{Scenario 2: Human Player Directing Agents}

\begin{bluebox}[title = Agent Steve's Environment Information]
\begin{lstlisting}
{
  "time": "day",
  "health": 1.2,
  "hunger": 17,
  "position": "(6, 66, 13)",
  "equipment": [
    "diamond_helmet",
    "diamond_chestplate",
    "diamond_leggings",
    "diamond_boots",
    "bow",
    null
  ],
  "inventory": {
    "arrow": 189,
    "golden_apple": 4,
    "bow": 1
  },
  "nearbyBlocks": [
    "end_stone"
  ],
  "nearbyEntities": "ender_dragon",
  "nearbyPlayers": [
    "Jeb"
  ]
}
\end{lstlisting}
\end{bluebox}

\begin{greenbox}[title = Human Player's Chat Message]
'Steve, you're in grave danger. Cease attack immediately and prepare to restore health!'
\end{greenbox}

\begin{bluebox}[title = Agent Steve's Planning Action]
\begin{lstlisting}
{
    "skill": "consumeItem(bot, \'golden_apple\', false)",
    "interrupt": true,
    "reason": "My health is critically low at 1.2, and I need to restore it immediately to continue fighting the ender_dragon effectively. I will follow the human player's advice and prioritize my health."
}
\end{lstlisting}
\end{bluebox}

\fi

\end{document}